\documentclass[11pt,a4paper]{article}

\usepackage[T1]{fontenc}
\usepackage[utf8]{inputenc}
\usepackage[margin=1in]{geometry}
\usepackage{amsmath,amssymb}
\usepackage{booktabs}
\usepackage{graphicx}
\usepackage{makecell}
\usepackage[numbers,sort&compress]{natbib}
\usepackage[hidelinks]{hyperref}
\usepackage{microtype}

\title{Acoustic-driven millimetric helical robot: ultrasonic synergistic manipulation in confined fluidic environments}
\author{%
Hanlin Wang\textsuperscript{1},
Xin Wang\textsuperscript{2},
Xinwei Wei\textsuperscript{3}, and
Jiaxu Liu\textsuperscript{4,*}\\
Le Wang\textsuperscript{5,6,*},
Shengze Cai\textsuperscript{1,*}, and
Chao Xu\textsuperscript{1,7,*}
}
\date{}

\begin{document}
\maketitle

\begin{center}
\begin{minipage}{0.96\textwidth}
\small
\textsuperscript{1}State Key Laboratory of Industrial Control Technology and the Institute of Cyber-Systems \& Control, Zhejiang University, Hangzhou 310027, Zhejiang, China\par
\textsuperscript{2}Huzhou Institute of Zhejiang University, Huzhou Key Laboratory of Robot System Integration and Intelligent Equipment, Huzhou 313099, Zhejiang, China\par
\textsuperscript{3}State Key Laboratory of Advanced Drug Delivery and Release Systems, College of Pharmaceutical Sciences, Zhejiang University, Hangzhou 310058, Zhejiang, China\par
\textsuperscript{4}School of Science, Huzhou Normal University, Huzhou 313000, Zhejiang, China\par
\textsuperscript{5}School of Intelligent Manufacturing and Elevator, Huzhou Vocational and Technical College, Huzhou 313099, Zhejiang, China\par
\textsuperscript{6}The Huzhou Key Laboratory of Robot System Integration and Intelligent Equipment, Huzhou 313099, Zhejiang, China\par
\textsuperscript{7}Zhejiang Provincial Engineering Research Center for Intelligent Mobile Unmanned Systems Technology and Huzhou Key Lab for Autonomous Systems, Huzhou Institute of Zhejiang University, Huzhou 313000, Zhejiang, China\par
\medskip
\textsuperscript{*}\textit{Corresponding authors:} jiaxuliu@zju.edu.cn (Jiaxu Liu); 2023019@huvtc.edu.cn (Le Wang); shengze\_cai@zju.edu.cn (Shengze Cai); cxu@zju.edu.cn (Chao Xu)\par
ORCID: Chao Xu, 0000-0002-2759-6364
\end{minipage}
\end{center}

\begin{center}
\footnotesize Author-prepared manuscript. For the version of record, see \emph{Ultrasonics}, DOI: \url{https://doi.org/10.1016/j.ultras.2026.108233}. This manuscript is shared under the Creative Commons Attribution-NonCommercial-NoDerivatives 4.0 International license (CC BY-NC-ND 4.0).
\end{center}

\begin{abstract}
Acoustic field–driven manipulation provides a non-contact and non-invasive strategy for controlling microscale and nanoscale objects, yet its extension to millimeter-scale robots was limited by insufficient propulsion efficiency in confined biological environments. Here, a coordinated multi-acoustic-field approach is introduced, which harnesses the synergistic action of acoustic radiation forces and acoustic streaming flows to enable controlled locomotion of millimeter-scale helical robots and enhance propulsion. Multiphysics simulations captured the dynamics of millimeter-scale helical robots under combined acoustic fields, and experimental validation demonstrated their locomotion capabilities, including planar navigation, inclined climbing, and vertical motion. Semi-autonomous navigation experiments further confirmed that ultrasonic synergy substantially improved maneuverability. In vitro tests in porcine venous vessels demonstrated that coordinated acoustic fields supported both unidirectional and reciprocating motion under biologically relevant confinement. These findings provide mechanistic insight into scaling acoustic micromanipulation to the millimetre regime and support biomedical applications requiring versatile and controllable robotic mobility.
\end{abstract}

\noindent\textbf{Keywords:} Millimeter-scale robot; Acoustic actuation; Synergistic propulsion

\section*{Highlights}
\begin{itemize}
\item A dual acoustic-field strategy, combining radiation forces and streaming flows, enables controlled locomotion of millimeter-scale helical robots in confined biological environments.
\item Versatile mobility—including planar, inclined, and vertical motions—allows semi-autonomous robots to execute complex path-following tasks.
\item In vitro vascular tests demonstrate bidirectional, sub-millimeter-precision motion, establishing a non-contact, biocompatible strategy for minimally invasive navigation.
\end{itemize}

\section{Introduction}

The past decade has witnessed paradigm-shifting advancements in medical microrobotics\cite{RN1}, revolutionizing minimally invasive interventions through sub-micrometer spatial precision and significantly reduced iatrogenic damage, thereby redefining targeted therapeutic and diagnostic approaches. Four principal actuation modalities have emerged as technological cornerstones: ultrasonic\cite{RN2,RN3,RN4,RN5,RN6,RN7,RN8}, magnetic\cite{RN9,RN10,RN11,RN12}, optical\cite{RN13,RN14}, and electrical\cite{RN15,RN16} systems, each demonstrating distinct performance characteristics that govern their translation potential. Optical manipulation platforms, employing either optical tweezers or photophoretic propulsion, achieve unparalleled nanometer-scale resolution through non-contact force delivery, establishing them as indispensable tools for single-cell manipulation and subcellular studies\cite{RN13}, yet face fundamental limitations including tissue penetration constraints due to scattering losses, non-negligible photothermal effects, and stringent refractive index matching requirements. The miniaturization of conventional electric-driven microrobotics is constrained by the precision limitations of fabrication processes and energy density bottlenecks. In contrast, piezoelectric ceramic materials demonstrate significant advantages due to their unique electromechanical coupling properties\cite{RN15}. Magnetic actuation demonstrates superior clinical relevance through deep-tissue penetration, real-time 3D navigation, and inherent biocompatibility of ferrite materials, yet contends with high energy demands, material property tradeoffs, and biological flow interference\cite{RN11}.

Acoustic microrobotics has emerged as a promising approach to overcoming the long-standing trade-off between penetration depth and microscale manipulation precision by exploiting the unique physics of acoustic wave–matter interactions. The technology's paramount advantage resides in acoustic radiation forces demonstrating favorable scaling laws with particle dimensions, permitting control across multiple length scales-from nanometer-sized drug carriers to millimeter-scale devices-while achieving exceptional tissue penetration depths of 10-15 cm at diagnostic ultrasound frequencies (1-10 MHz) \cite{RN7,RN8}. Contemporary research has uncovered significant performance limitations in confined microfluidic environments, where conventional acoustic microrobot designs experience substantial velocity attenuation in submillimeter channels due to pronounced boundary interactions\cite{RN12}. These operational constraints originate from intrinsic physical limitations of existing actuation modalities: magnetic systems exhibit exponential force decay with distance\cite{RN10,RN11}, while optical methods face insurmountable penetration depth restrictions\cite{RN13}. Clinical applications such as positional accuracy in dynamic physiological flows necessitate breakthrough capabilities \cite{RN1}, while industrial microtransport scenarios demand reliable navigation through complex microchannel networks \cite{RN12}. These multifaceted challenges highlight the critical necessity for novel propulsion architectures that overcome current physical limitations through advanced control paradigms and innovative actuation mechanisms.

Current research on acoustic microrobots identifies three critical knowledge barriers preventing clinical and industrial translation\cite{RN17,RN18,RN19,RN20,RN21}. First, no comprehensive locomotion framework to predict coupled rotational and translational motions of microrobots in confined flows, particularly regarding the synergistic interplay among acoustic streaming(the steady fluid flow induced by viscous attenuation of acoustic waves), radiation forces, self-rotation dynamics, and boundary friction effects under ultrasonic excitation. Second, the complex interplay between acoustic radiation forces and acoustic streaming-induced locomotion remains poorly quantified. Third, at the submillimeter scale, the compounded effects of Brownian motion, fluid drag, and signal attenuation collectively degrade robot’s stability, precision, and controllability\cite{RN22,RN36}, despite its high clinical relevance for targeted drug delivery and minimally invasive surgery\cite{RN23,RN24}. These three critical limitations underscore the urgent need for integrated multiphysics coupling dynamics modeling and cross-scale control theories in millimeter-scale robotics research.

An integrated approach combining theory, simulation, and experiment enables acoustic millimetric helical robots to achieve controlled locomotion and expands their operational capabilities. A fundamental framework clarifies their propulsion mechanisms through rigorous analysis of driving principles and simulations of spiral-structure acoustics. Multiphysics computational models successfully solve the complex fluid-structure interaction (FSI) and acoustic wave propagation problems in aqueous environments. Through this theoretical investigation, it is conclusively demonstrated that the robotic propulsion arises from four synergistic mechanisms: acoustic radiation force propulsion, acoustic streaming propulsion, robotic self-rotation effects, and boundary friction effects. The experimental validation begins with sophisticated particle image velocimetry (PIV) measurements \cite{RN25,RN26,RN27,RN28} that quantitatively confirm the remarkable consistency between the observed ultrasonic streaming patterns in fluidic environments and our simulation predictions. Furthermore, the significant enhancement of acoustic streaming propulsion is revealed when multiple piezoelectric transducer (PZTs) drivers are employed through carefully designed interference patterns.

Building upon insights, systematic structural optimization of the spiral microrobots is conducted by establishing quantitative relationships between various geometric parameters (including pitch angle, helix diameter, and surface roughness) and propulsion performance. The motion capabilities of our optimized microrobots are then thoroughly characterized across multiple challenging scenarios: two-dimensional planar motion demonstrates controllability, while inclined-surface climbing reveals remarkable adaptability, and vertical surface adhesion showcases stability against gravitational forces. These comprehensive locomotion studies provide unprecedented insights into the microrobots' performance in realistic environments. Our semi-autonomous navigation control system further demonstrates the platform's excellent controllability and robust operation under varying conditions. The clinical potential of this technology is conclusively validated through pioneering in vitro experiments using porcine venous vasculature, where the microrobots exhibit superior unidirectional motion performance and stable reciprocating movement patterns at biological tissue interfaces. These vascular navigation experiments represent a critical advancement in minimally invasive medical robotics by addressing the long-standing challenge of controlled propulsion in complex biological environments. The integration of theoretical modeling, structural optimization, motion characterization, and biological validation in this work advances the development of acoustic microrobots with enhanced precision, reliability, and clinical relevance.

\section{Method}
In this section, the theoretical principles underlying the propulsion of millimeter-scale helical robots under acoustic excitation are presented. The robot experiences synergistic actuation through acoustic streaming and acoustic radiation forces. Acoustic streaming arises from nonlinear energy conversion, where acoustic energy is transformed into fluid kinetic energy through viscous dissipation and boundary layer effects. Concurrently, the acoustic radiation force emerges from momentum transfer between the acoustic field and the robot, enabling manipulation along the pressure gradient vectors.

The following analysis investigates the actuation mechanism of the helical millirobot using two primary approaches: (a)Time-domain flow field and frequency-domain acoustic field simulations within the water tank environment to capture fluid dynamics and the contribution of acoustic streaming effects. (b)Acoustics analysis of the helical structure to resolve pressure propagation and acoustic radiation effects.

\subsection{Synergistic propulsion mechanism}

In this complex multiphysical system, the  helical robot experiences synergistic actuation through simultaneous interaction with acoustic streaming\cite{RN29,RN30,RN31,RN32} and radiation forces\cite{RN33,RN34,RN35} under acoustic field excitation. Acoustic streaming manifests as directional fluid motion arising from nonlinear energy conversion where acoustic energy is transformed into fluid kinetic energy through viscous dissipation and boundary layer effects. Concurrently, acoustic radiation force emerges as a net momentum transfer between the acoustic field and solid objects, enabling manipulation of suspended robotic components along pressure gradient vectors. This combined mechanism demonstrates remarkable advantages in low-disturbance, high-maneuverability scenarios by leveraging fluid-structure-acoustic interactions. The observed synergy between these two phenomena opens new avenues for micro-robotic propulsion and bioinspired locomotion.

\subsubsection{Acoustic streaming in water tank}
Acoustic streaming in the bulk region is governed by the Eckart force, which arises from the radiation pressure of acoustic waves and drives large-scale vortical structures. This nonlinear phenomenon occurs when the second-order acoustic streaming velocity field balances viscous dissipation, establishing a dynamic equilibrium. Through dimensional analysis, the approximate expression for the acoustic streaming velocity is derived:

\begin{equation}
U_s \approx \frac{\alpha c_0 \rho_0 U_1^2 L^2}{\mu}
\end{equation}

Where $\alpha$ is the attenuation coefficient, $C_0$, $\rho_0$ and $\mu$ represent the speed of sound, density and dynamic viscosity in the water, $U_1$ is the particle velocity amplitude with $L$ the hydraulic diameter of the glass channel.

\subsubsection{Acoustic radiation force for helical robot}
Driving by wall-bonded PZT emitters around 45–50 kHz, so the robot(diameter $D=2.5\mathrm{~mm}$ , pitch  
$p=1.5\mathrm{~mm}$, length $L=6\mathrm{~mm}$  ) lies in the Rayleigh subwavelength regime $k a=\frac{2 \pi}{\lambda} a \approx 0.21 \ll 1$ with $a \sim \frac{D}{2}$ .
Based on the Rayleigh long wavelength approximation then have: 
\begin{equation}F_x =\left(\sum_{i=1}^N \Delta V_i\right) 2 k E_{a c} \Phi \sin \left(2 k x_i\right) \approx 2 V k E_{a c} \Phi \sin \left(2 k x_c\right)
\end{equation}

where $F_x$ is the acoustic radiation force along the x-direction, $ V_i$ denotes the induced velocity component, and $E_{a c}$ represents the acoustic energy density.

In the Rayleigh regime, the potential acquires an orientation dependence through the anisotropy of the dipole response, giving:

\begin{equation}\mathbf{N}_{\mathrm{rad}}=-\frac{\partial U}{\partial \Theta} \approx-\frac{1}{2} V \rho_0 \Delta f_1\left(\left\langle\mathbf{u}_{1, \|}^2\right\rangle-\left\langle\mathbf{u}_{1, \perp}^2\right\rangle\right) \sin (2 \theta) \hat{\mathbf{n}}
\end{equation}

wher…6336 tokens truncated…hile leftward steering required 2–3s due to frequency-dependent impedance stabilization.

The anisotropy between rightward and leftward steering originates from chiral interactions between the helical fins and asymmetric acoustic standing-wave patterns. Near bends, clearance between the fins and channel walls (~0.2 ± 0.05 mm) created localized acoustic impedance discontinuities, breaking flow symmetry and initiating reorientation. Real-time phase modulation subsequently reconfigured nodal distributions, reversing thrust vectors within 1.5 s. Enhanced propulsion efficiency was attributed to three synergistic mechanisms: (a) controlled thrust vectoring through bimorph phase-differential actuation, (b) acoustic streaming reinforcement along helical grooves, and (c) hydrodynamic lubrication layers that reduced wall friction during spiral rotation. These experiments demonstrate the feasibility of tunable microrobotic navigation in tortuous fluidic networks. By linking phase-modulation parameters to real-time environmental feedback, this approach enables semi-autonomous navigation and supports future biomedical applications such as targeted thrombus clearance and exploration of branched vascular systems.

\subsection{Vitro validation in porcine vascular models}

Helical microrobot navigation in biological environments was experimentally realized using freshly excised porcine venous vessels to assess propulsion stability under heterogeneous acoustic transmission. The experimental platform consisted of a four-sided PZT-actuated water tank (50 kHz, independently addressable), within which glass microtubes and vascular specimens were sequentially mounted (Fig. 7a-b). This configuration enabled direct comparison between rigid glass conduits and compliant venous tissue, and provided a platform for evaluating locomotion performance across medium transitions. The robot employed in these studies had a diameter of 2.5 mm and a pitch of 1.5 mm.

Two representative propulsion modes were designed to validate performance. In the L1 linear propulsion mode (Fig.7d), the robot traversed glass tubing and continued directly into porcine veins under the application of a unidirectional acoustic field (AW1). Fig. 7c shows the displacement–time profile, where the robot reached a baseline velocity of 19.1 mm/s in glass tubing and, after crossing the glass–vasculature interface at $t=1s$, maintained 17.4 ± 0.5 mm/s inside the vein. Notably, by $t=5s$, it had completed over 80 mm of heterogeneous conduit traversal, preserving $91 \%$ of its baseline velocity despite increased wall friction and viscoelastic damping. This demonstrated the stabilizing role of spiral groove–mediated acoustic streaming and adaptive friction modulation in sustaining forward propulsion across media boundaries.

In the L2 reciprocating propulsion mode (Fig.7e), the robot advanced through glass tubing and entered the porcine vein under AW1 excitation, then reversed direction under a counter-propagating acoustic field (AW2). During the forward phase ($t=0-4s$), the robot reached 13.0 mm/s inside the vein. At $t=4s$, AW2 was applied, inverting the radiation force vector by 180°. Under combined effects of vascular wall friction and the opposing acoustic thrust, the robot decelerated to rest and subsequently accelerated retrogradely, reaching –9.7 ± 0.5 mm/s. By $t=9.5s$, it had returned to its starting position, completing a full bidirectional cycle within the vascular lumen. The displacement–time trajectory (Fig.7c) illustrates this sequence, highlighting both the deceleration plateau and the transition to retrograde acceleration. Although retrograde propulsion was $25 \%$ less efficient than forward locomotion due to viscoelastic damping, real-time impedance matching (1 kHz sampling) dynamically tuned the excitation frequency between 45–55 kHz, maintaining ±0.3 mm positional precision throughout bidirectional operation.

These results show that helical microrobots can traverse biological conduits with stable forward propulsion and undergo reversible navigation in compliant vascular environments. The combination of spiral geometry and phase-controlled acoustic excitation supports both sustained linear locomotion and retrograde return without mechanical reconfiguration. Such controllable locomotion supports the development of endovascular microrobots for applications including targeted drug delivery, thrombus clearance, and multi-site tissue sampling in complex vascular networks.

To place these biological validation experiments in an appropriate translational context, acoustic safety considerations were discussed with reference to commonly used diagnostic-ultrasound metrics, including derated spatial-peak temporal-average intensity (ISPTA.3) and mechanical index (MI). For peripheral vessel applications, commonly cited reference limits are ISPTA.3 < 720 mW/cm² and MI < 1.9. In the present study, no visible cavitation, bubble formation, or macroscopic tissue damage was observed during ex vivo exposure. However, because these experiments were performed as a proof-of-concept demonstration, a full safety evaluation—including quantitative assessment of acoustic intensity, tissue attenuation, local temperature rise, and long-term bioeffects—will be required before clinical translation.

The ex vivo porcine vein experiments were conducted under static conditions and were intended to evaluate the feasibility of robot locomotion in a confined biological structure rather than to reproduce realistic in vivo hemodynamic environments. As such, the present results do not account for background blood flow, pulsatility, wall shear stress, or blood–robot interactions, all of which may substantially influence propulsion and navigation performance in vivo. The biological validation presented here should therefore be interpreted as a proof-of-concept demonstration under structurally relevant but flow-free conditions. Future studies will be needed to assess robot behavior under physiologically relevant flow fields, including both steady and pulsatile background flow, and thereby more fully evaluate the potential of the proposed strategy for intravascular applications.

\section{Conclusion}
This work demonstrates that coordinated acoustic fields can overcome longstanding barriers to propulsion efficiency in millimeter-scale robotic systems, enabling controlled locomotion in both artificial and biological environments. By integrating multiphysics simulations with systematic experimental validation, it is established how helical morphology, combined with spatiotemporally modulated acoustic fields, produces controllable translational and rotational dynamics. From planar navigation and inclined climbing to vertical ascent, the experiments confirmed that propulsion performance derives from the synergistic action of acoustic radiation forces and streaming flows, whose relative contributions can be tuned through transducer activation sequences and phase differentials. Semi-autonomous navigation in tortuous S and Y-shaped channels further illustrated how programmable acoustic fields enable directional switching and path selection, advancing microrobotic maneuverability beyond simple forward propulsion.

Biological validation within porcine venous vessels revealed that millimeter-scale helical robots sustain robust propulsion across heterogeneous media. Beyond confirming feasibility, these results highlight the broader implications of scaling acoustic micromanipulation into the millimeter regime. First, they extend the reach of non-contact acoustic actuation from microscale droplets and particles to functional devices capable of navigating vascular-sized conduits. Second, they demonstrate that multimodal locomotion can be achieved without mechanical reconfiguration, relying instead on adaptive acoustic field modulation. Third, they point toward strategies for overcoming heterogeneous attenuation in tissue by integrating real-time phase control and impedance matching. Nevertheless, several challenges remain before translation to in vivo deployment, including acoustic safety at clinically relevant voltages, real-time imaging for closed-loop control, and long-term biocompatibility of robotic materials.\\

\textbf{Experimental Setup and Instrumentation}

The millimeter-scale helical robot executes swimming motions in a $100 \times 100\times$ 100 mm glass aquarium, with four square piezoelectric transducers bonded to the exterior walls of each side.

The system was configured to quantify the acoustic streaming dynamics and robotic propulsion mechanisms in a customized water tank. The setup included:Dual-pulsed 532 nm Nd laser (200 mJ/pulse, 10 Hz) with a cylindrical lens to generate a 1 mm-thick light sheet. High-speed CMOS camera (500 fps, 2048 × 2048 pixels) synchronized with laser pulses via a programmable timing unit. Fluorescent polystyrene particles (10 $\mu \mathrm{m}$ diameter, $\rho=1.05 \mathrm{~g} / \mathrm{cm}^3$) seeded at $0.02 \%$ volume concentration. Four square piezoelectric transducers (PZTs; 50 × 50 × 1 mm³, resonant frequency 50 kHz), driven by independent function generators (0–$150\mathrm{~V}_{\mathrm{p}-\mathrm{p}}$, 45–55 kHz).\\

\textbf{PIV Protocol and Data Acquisition}

Spatial calibration used a dual-level target (0.5 mm grid) to achieve 20 $\mu \mathrm{m}$/pixel resolution. For each experimental condition, 500 image pairs were acquired with optimal pulse separation ($\Delta \mathrm{t}=200\mu\mathrm{~s}$) determined through preliminary tests to maintain particle displacements below $25 \%$ of the interrogation window size. Cross-correlation analysis was conducted using 32 × 32 pixel interrogation windows with $50 \%$ overlap, yielding vector fields with 0.5 mm spatial resolution. 

Three distinct activation modes were systematically investigated: Mode 1 employed single-PZT excitation (PZT1 at 50 kHz,  $100\mathrm{~V}_{\mathrm{p}-\mathrm{p}}$) to establish baseline streaming patterns, Mode 2 implemented dual-PZT in-phase actuation (PZT1+PZT2, $\Delta\varphi=0^{\circ}$) for constructive interference, Mode 3 utilized triple-PZT phased activation (PZT1+PZT2+PZT3, $\Delta \varphi=90^{\circ}$) to generate enhanced vorticity fields. 
Post-processing involved median filtering ($\pm 3 \sigma$ threshold) to remove spurious vectors followed by Gaussian smoothing (3 × 3 pixel kernel) to reduce measurement noise.\\

\textbf{Biological Environment Validation}

The porcine vascular segments obtained from commercial markets underwent rigorous validation to simulate the biological environment for millimeter-scale helical robot navigation.
Prior to experimentation, the samples were systematically characterized to confirm their structural integrity and physiological relevance, 
including histological analysis to evaluate endothelial layer preservation and mechanical testing to assess compliance with typical porcine vascular parameters.

In biological validation experiments using porcine venous vessels, a high-speed camera tracked robot motion across tissue–glass interfaces at 500 fps, demonstrating the applicability of the technique in complex biomedical settings. Three independent trials were conducted for each condition (n = 15 datasets in total) to assess the repeatability of robot motion under identical acoustic actuation, rather than to imply fully reliable first-attempt control. The results indicate consistent open-loop navigation in confined vascular environments, while further improvements in real-time feedback and closed-loop control will be required to achieve higher first-run trajectory accuracy. All experiments were conducted in accordance with institutional ethical guidelines for biological specimen handling.






\begin{figure}[h]
	\centering
	\includegraphics[width=\textwidth,height=0.82\textheight,keepaspectratio]{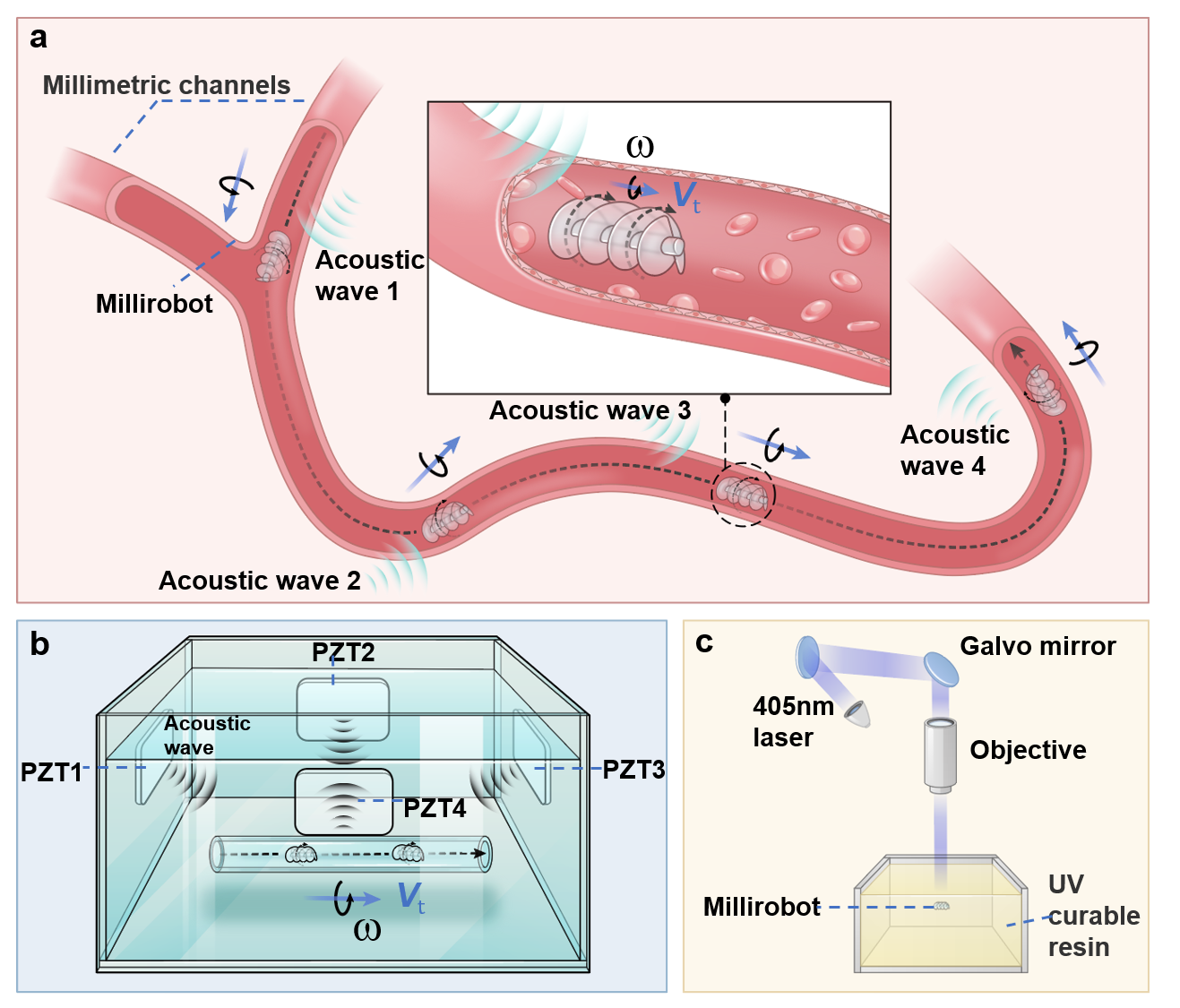}
	\caption{a) Schematic diagram of locomotion of a millimetric helical robot in blood vessels. b) The multi-sound field collaborative driving of the helical robot moving in confined pipeline. c)  3D printing manufacturing of the helical robot.}
	\label{fig:1}
\end{figure}

\begin{figure}[h]
	\centering
	\includegraphics[width=\textwidth,height=0.82\textheight,keepaspectratio]{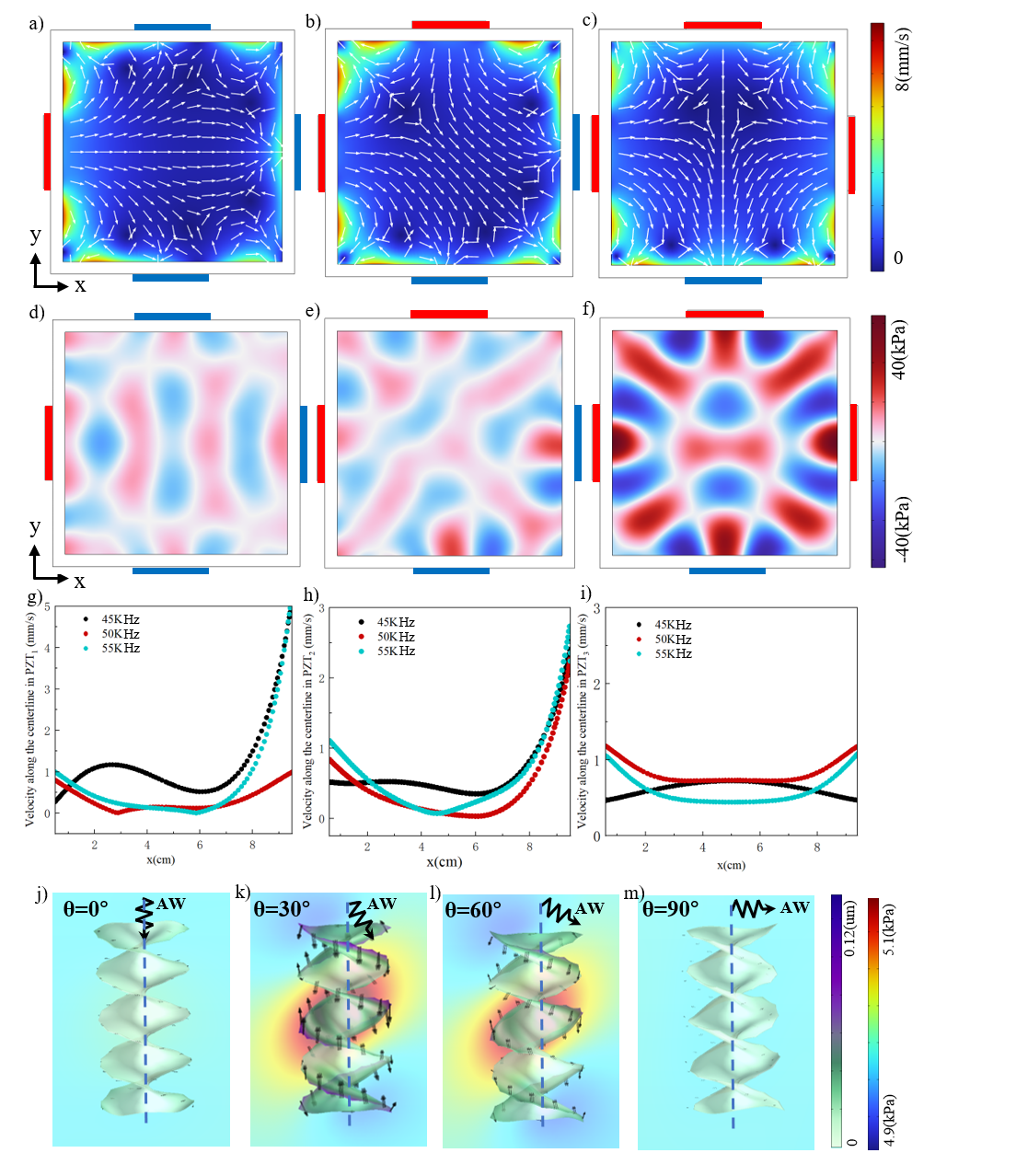}
	\caption{Flow field and acoustic pressure simulations in a water tank with robotic actuation. a–c) Flow field dynamics under actuation by (a) single-PZT, (b) dual-PZT, and (c) triple-PZT configurations. Activated PZTs (red) and inactive PZTs (blue) are distributed on the tank walls. d–f) Corresponding acoustic pressure distributions for the three actuation states in a–c. g–i) Axial velocity distributions along the tank centerline under ultrasonic frequencies of 45 kHz, 50 kHz, and 55 kHz for (g) single-PZT, (h) dual-PZT, and (i) triple-PZT actuations. j–m) Simulated acoustic pressure (color gradient) and displacement fields (vector arrows) acting on a helical robot, with the acoustic wave direction oriented at anglesof $\theta=0^{\circ}, 30^{\circ}, 60^{\circ}$, and $90^{\circ}$ relative to the robot’s central axis.}
	\label{fig:2}
\end{figure}

\begin{figure}[h]
	\centering
	\includegraphics[width=\textwidth,height=0.82\textheight,keepaspectratio]{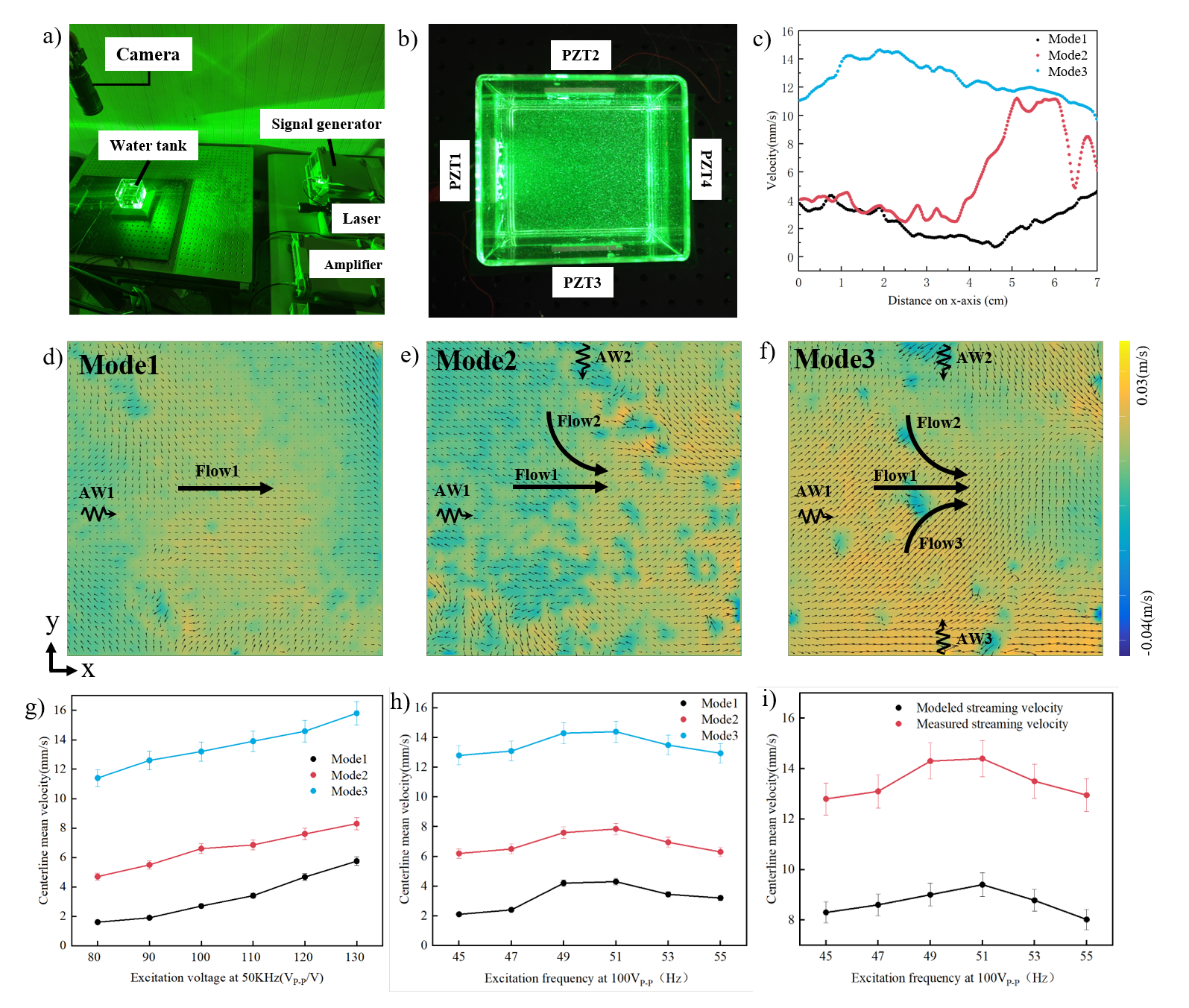}
	\caption{PIV experimental validation of the flow field within the water tank.(a) PIV experimental platform.(b) Local distribution of multiple PZT ceramics on the tank wall.(c) Spatial velocity field along the axial direction of PZT1 and PZT4. The black line denotes the centerline velocity distribution under mode 1, the red line corresponds to mode 2, and the blue line corresponds to mode 3. (d)–(f) Flow-field distributions obtained under the three activation modes.(g) Measured centerline streaming velocity as a function of input voltage at 50 kHz for modes 1–3.(h) Measured centerline streaming velocity as a function of driving frequency at 100 Vpp for modes 1–3.(i)Comparison between modeled and measured centerline streaming velocities under mode 3. Both modeled and measured results show a maximum near 49–51 kHz, while the measured velocities are consistently higher than the modeled predictions, likely owing to simplified acoustic boundary conditions and incomplete representation of the transducer–tank interaction in the numerical model.}
	\label{fig:3}
\end{figure}

\begin{figure}[h]
	\centering
	\includegraphics[width=\textwidth,height=0.82\textheight,keepaspectratio]{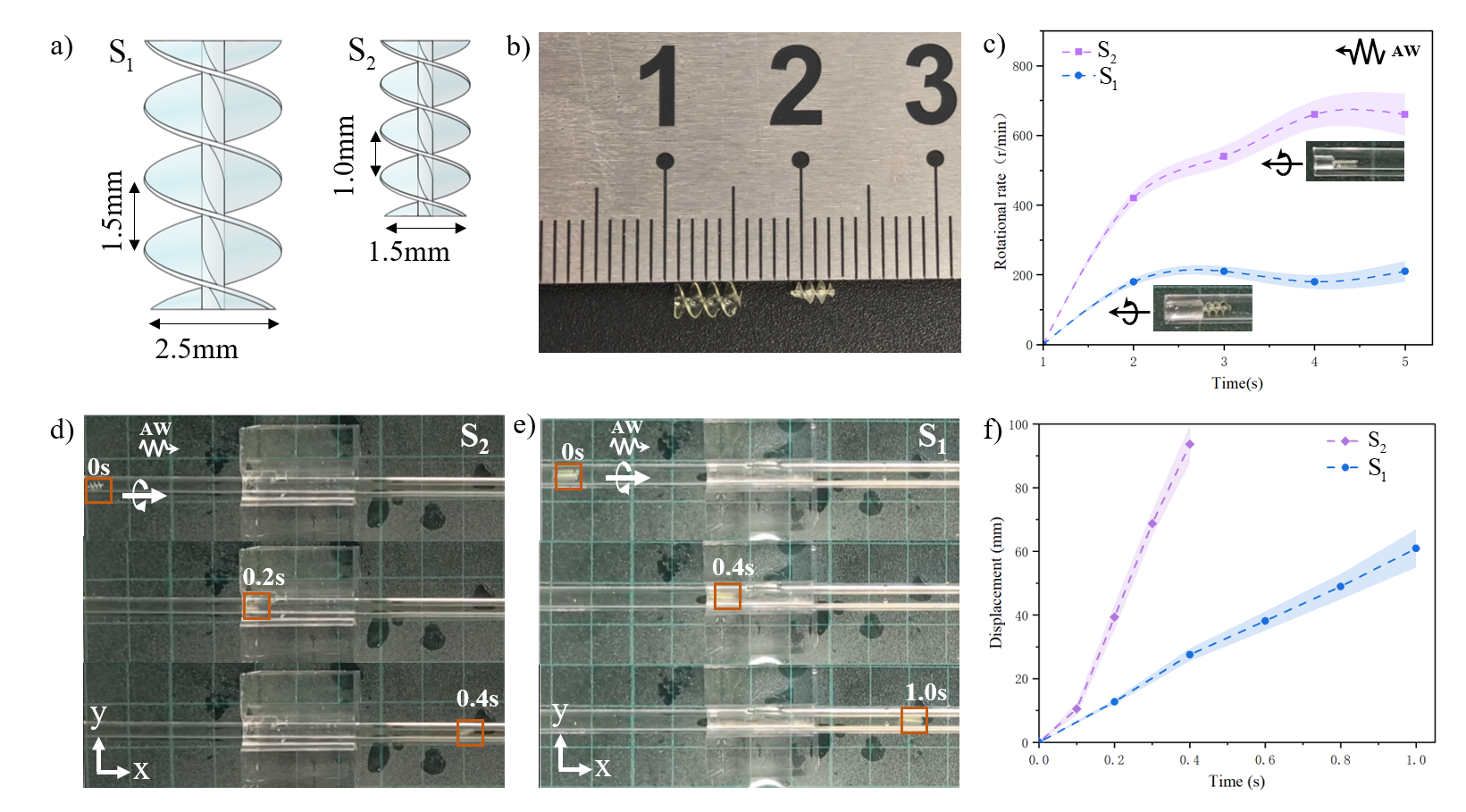}
	\caption{Performance comparison of helical robots with varying dimensions.
a) Schematic illustration of robot dimensions. b) Scale comparison of the two helical robots. c) Time-dependent rotational speed curves for${S}_1$ and ${S}_2$  robots under ultrasonic actuation. d–e) Horizontal motion trajectories of (d) ${S}_2$  and (e)  ${S}_1$  robots. f) Displacement-time curves for both robots under ultrasonic excitation. The results demonstrate that smaller robots exhibit enhanced translational velocity.}
	\label{fig:4}
\end{figure}

\begin{figure}[h]
	\centering
	\includegraphics[width=\textwidth,height=0.82\textheight,keepaspectratio]{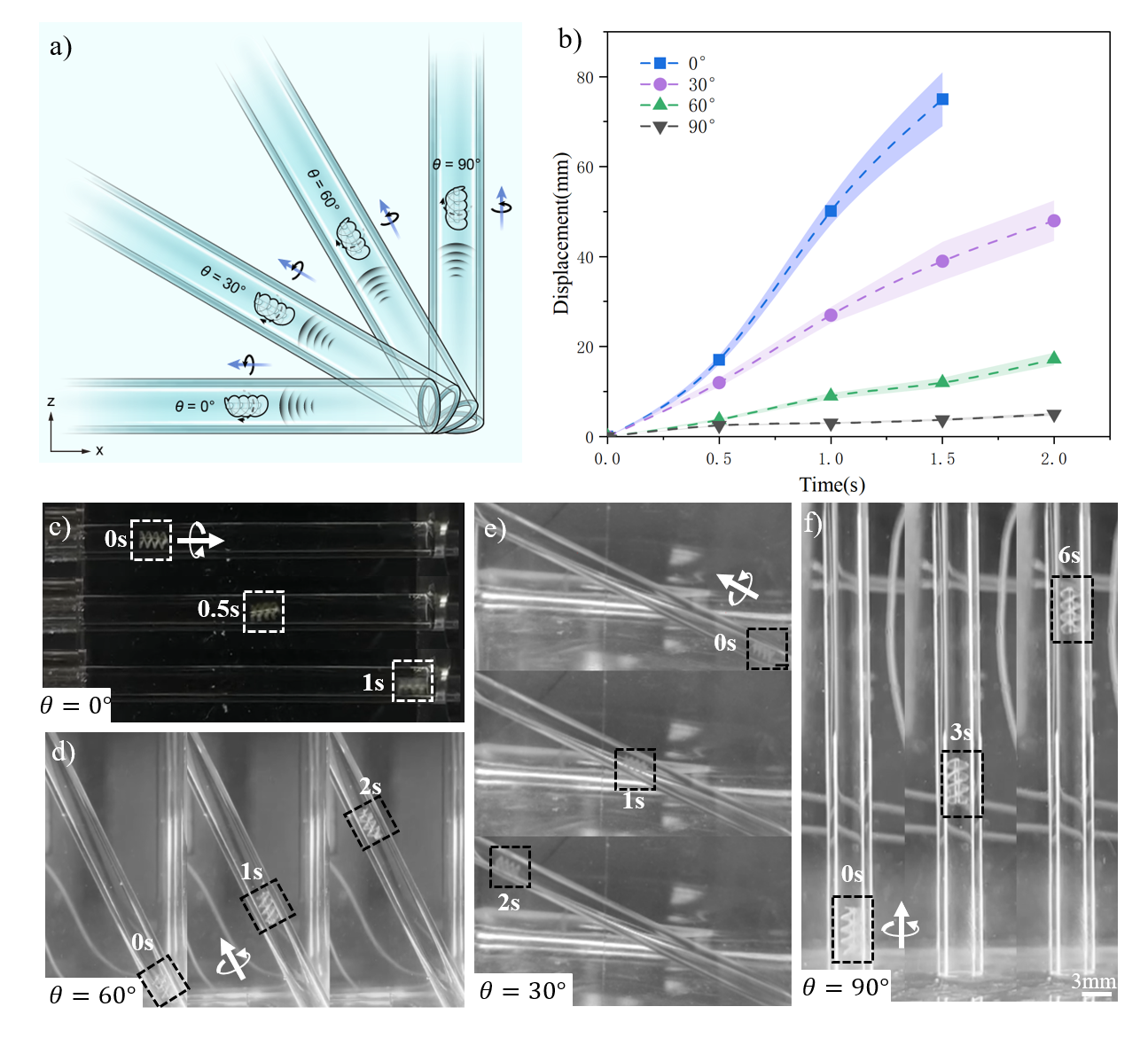}
	\caption{3D manipulation of microrobots in microfluidic channels.
a) Schematic illustrating helical microrobot navigation at varying tilt angles within the channel. b) Comparative displacement-time curves for different tilt angles. c-f) Composite images showing the microrobot's horizontal motion and vertical propulsion at (c) 0°, (e) 30°, (d) 60°, and (f) 90° tilt angles.}
	\label{fig:5}
\end{figure}

\begin{figure}[h]
	\centering
	\includegraphics[width=\textwidth,height=0.82\textheight,keepaspectratio]{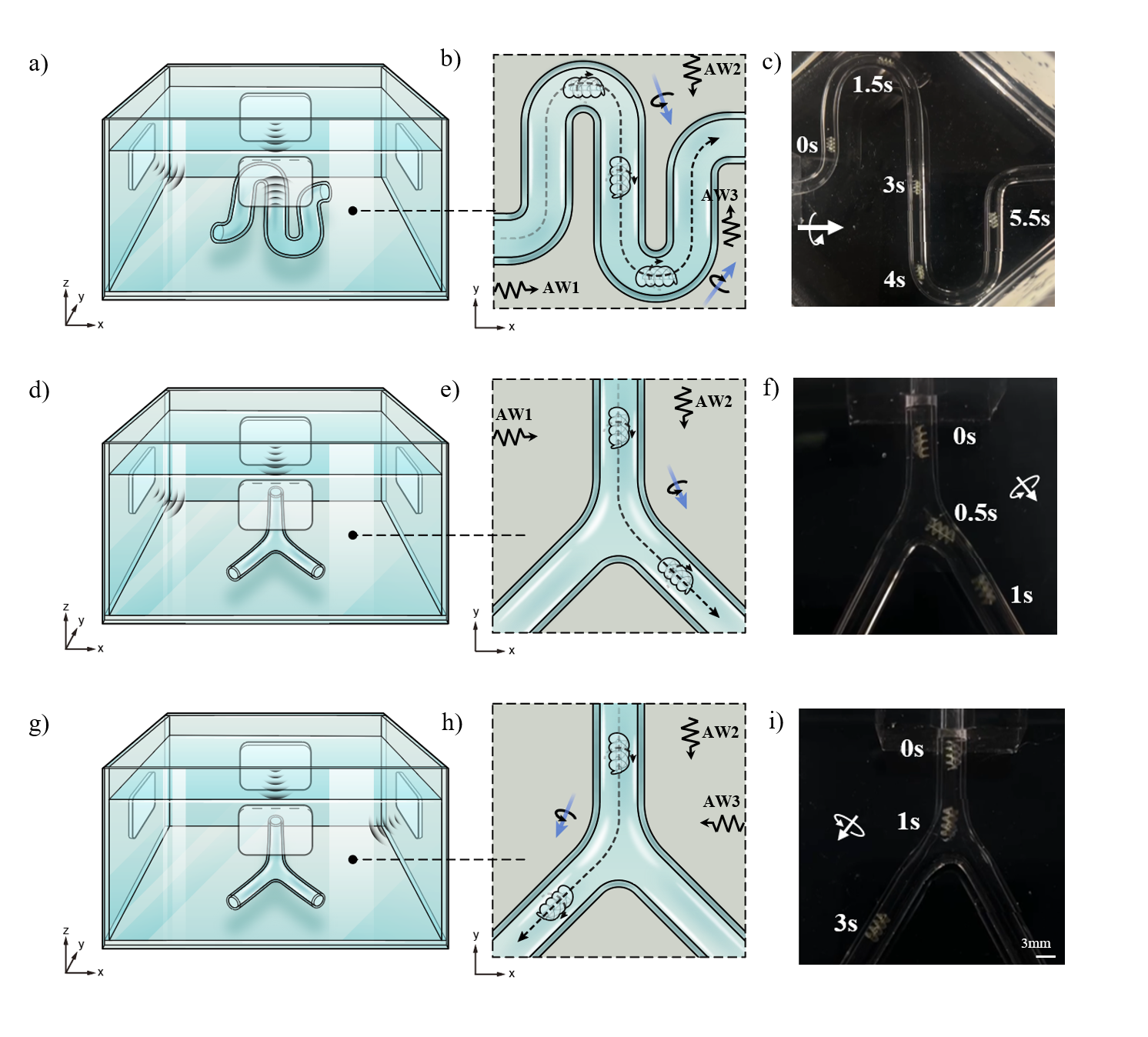}
	\caption{ Microrobotic navigation in complex microfluidic channel systems.
a–c) Helical microrobot propulsion in an S-shaped channel under triple-PZT actuation. d–f) Leftward navigation in a Y-shaped channel via alternating dual-PZT actuation. g–i) Rightward navigation in the same Y-shaped channel configuration under dual-PZT switching.}
	\label{fig:6}
\end{figure}

\begin{figure}[h]
	\centering
	\includegraphics[width=\textwidth,height=0.82\textheight,keepaspectratio]{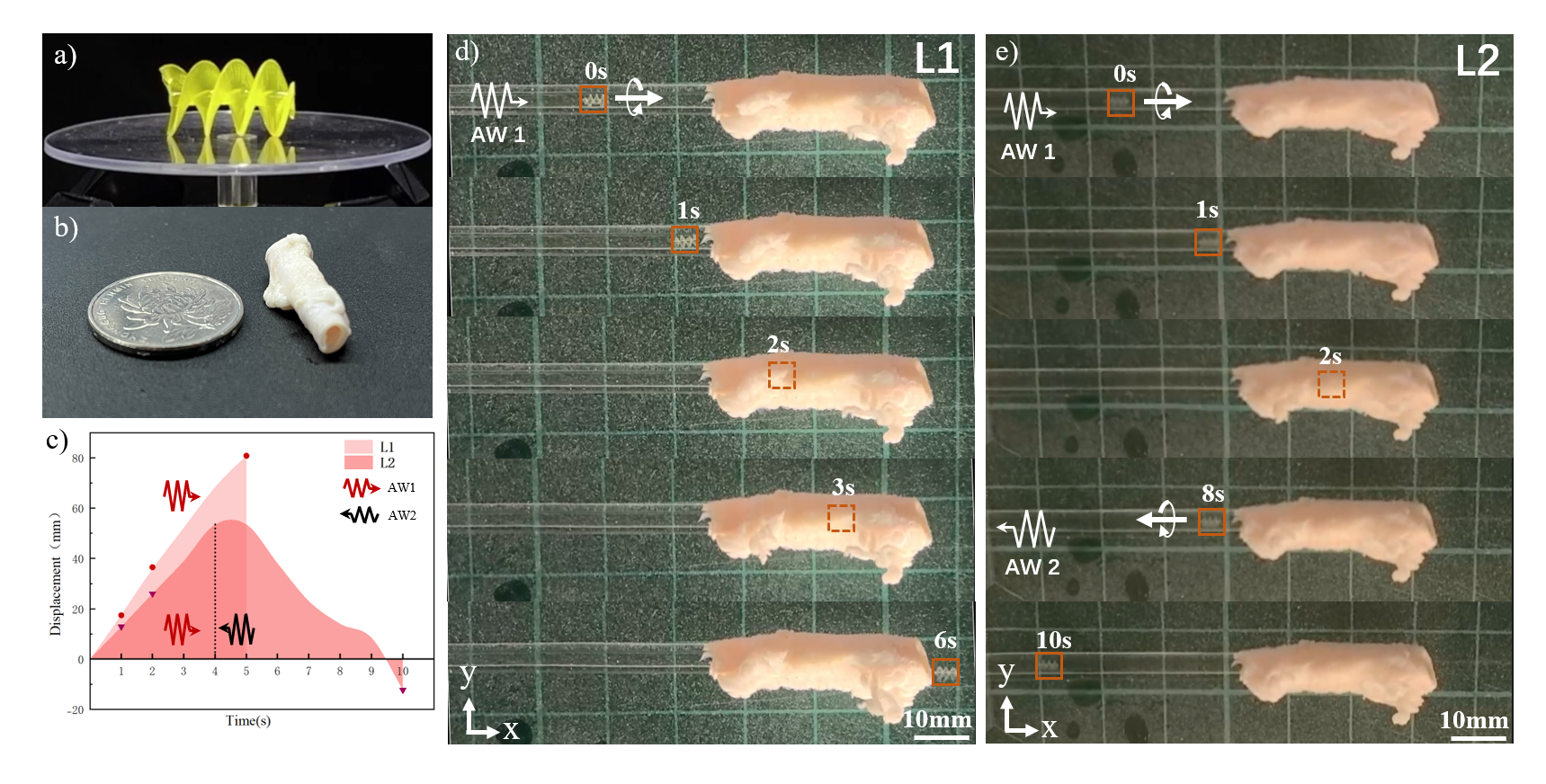}
	\caption{Microrobotic manipulation in ex vivo porcine vascular systems.
a) 3D-printed helical microrobot prototype. b) Fresh porcine vein segment with 3 mm inner diameter. c) Displacement-time profile under ultrasonic actuation: L1 (unidirectional translation) represents vascular entry/exit, while L2 (reciprocating motion) demonstrates return-to-origin via acoustic field reversal (AW1/AW2 with 180° phase difference). d-f) Experimental validation of (d) L1 task execution and (e) L2 task performance.}
	\label{fig:7}
\end{figure}

\begin{table}[htbp]
\centering
\setlength{\tabcolsep}{1pt}
\caption{Comparison of acoustically driven helical robots reported in the literature and in this work.}
\label{tab:comparison}

\resizebox{\textwidth}{!}{%
\begin{tabular}{lll}
\toprule
Parameter & Deng et al\cite{RN7}. & Wang et al. (This work) \\
\midrule
Size & Length 350 $\mu$m, diameter 100 $\mu$m & Length 6 mm, diameter 2.5 mm \\
Material & Polymeric material & Epoxy resin \\
Aspect ratio & 3.5 & 2.4 \\
Translational speed & 100 $\mu$m/s & 15 mm/s \\
Rotation speed & $\sim$100 RPM & $\sim$660 RPM \\
Driving frequency & 10--20 kHz & 45--55 kHz \\
Driving voltage & 20 V$_{pp}$ & 80--150 V$_{pp}$ \\
Boundary condition & Planar rectangular cavity; 500 $\mu$m microchannel  & Open rectangular fluid chamber; 
3 mm microchannel  \\
Actuation method & Single piezoelectric transducer & Piezoelectric ceramic array \\
Acoustic source & Buzzer & Piezoelectric ceramics \\
Scale & Microscale & Millimeter scale \\
\bottomrule
\end{tabular}%
}
\end{table}

\clearpage

\section*{CRediT authorship contribution statement}
\textbf{Hanlin Wang}: Writing - review \& editing, Writing - original draft, Visualization, Validation, Methodology, Investigation, Data curation, Conceptualization.
\textbf{Le Wang}: Writing - review \& editing, Validation, Methodology, Conceptualization.
\textbf{Xin Wang}: Validation, Methodology.
\textbf{Jiaxu Liu}: Validation, Methodology, Data curation.
\textbf{Xinwei Wei}: Validation, Methodology, Conceptualization.
\textbf{Shengze Cai}: Writing - review \& editing, Validation, Methodology, Conceptualization.
\textbf{Chao Xu}: Writing - review \& editing, Supervision, Resources, Project administration, Methodology, Funding acquisition.

\section*{Funding}
This document is the results of the research project funded by the Zhejiang Provincial Natural Science Foundation of China (Grant No. LZ24F030003),the Huzhou Public Welfare Application Research Project (2024GZ26) and General Scientific Research Projects of Zhejiang Provincial Department of Education (Grant No. Y202352861).

\bibliographystyle{unsrtnat}
\bibliography{references}

\end{document}